\documentclass[journal]{IEEEtran}
\usepackage{bibentry}
\usepackage{xcolor,soul,framed} 

\colorlet{shadecolor}{yellow}
\usepackage[pdftex]{graphicx}
\graphicspath{{../pdf/}{../jpeg/}}
\DeclareGraphicsExtensions{.pdf,.jpeg,.png}

\usepackage{amsmath}
\usepackage{amssymb}
\usepackage{array}
\usepackage{mdwmath}
\usepackage{mdwtab}
\usepackage{eqparbox}
\usepackage{url}
\usepackage{multirow}

\usepackage{algorithmic}
\usepackage{algorithm}

\usepackage{hyperref}
\hypersetup{hidelinks,
	colorlinks=true,
	allcolors=black,
	pdfstartview=Fit,
	breaklinks=true}
 
\begin{document}
\bstctlcite{IEEEexample:BSTcontrol}
\title{Beyond Pixel-Wise Supervision for Medical Image Segmentation: From Traditional Models to Foundation Models}
\author{Yuyan Shi, Jialu Ma, Jin Yang, Shasha Wang and Yichi Zhang
	\thanks{Y. Shi and S. Wang are with the Network and Data Center, Northwest University, Xi’an, China.}
    \thanks{J. Ma is with the School of Software Engineering and Systems, Lappeenranta University of Technology, Lahti, Finland, and Hebei University of Technology, Tianjin, China.}
	\thanks{J. Yang is with the Department of Radiology, Washington University in St.Louis, St. Louis, MO, USA.}
	\thanks{Y. Zhang is with the School of Data Science, Fudan University, Shanghai, China.}
}
\maketitle

\begin{abstract}
Medical image segmentation plays an important role in many image-guided clinical approaches.
However, existing segmentation algorithms mostly rely on the availability of fully annotated images with pixel-wise annotations for training, which can be both labor-intensive and expertise-demanding, especially in the medical imaging domain where only experts can provide reliable and accurate annotations.
To alleviate this challenge, there has been a growing focus on developing segmentation methods that can train deep models with weak annotations, such as image-level, bounding boxes, scribbles, and points. 
The emergence of vision foundation models, notably the Segment Anything Model (SAM), has introduced innovative capabilities for segmentation tasks using weak annotations for promptable segmentation enabled by large-scale pre-training.
Adopting foundation models together with traditional learning methods has increasingly gained recent interest research community and shown potential for real-world applications.
In this paper, we present a comprehensive survey of recent progress on annotation-efficient learning for medical image segmentation utilizing weak annotations before and in the era of foundation models. Furthermore, we analyze and discuss several challenges of existing approaches, which we believe will provide valuable guidance for shaping the trajectory of foundational models to further advance the field of medical image segmentation.
\end{abstract}

\begin{IEEEkeywords}
Medical Image Segmentation, Annotation-Efficient Learning, Weakly Supervised Learning, Foundation Models, Survey.\\ \\
\end{IEEEkeywords}

\IEEEpeerreviewmaketitle

\section{Introduction}
\IEEEPARstart{M}{edical} image segmentation aims to delineate the interested anatomical structures like organs and tumors from the original images by labeling each pixel into a certain class, which is one of the most representative and comprehensive research topics in the community of medical image analysis \cite{MIA2017survey,Lynch2018NewMT}. 
Accurate segmentation can provide reliable volumetric and shape information of target structures and assist in many further clinical applications like disease diagnosis, quantitative analysis, and surgical planning \cite{AbdomenCT-1K,qureshi2023medical}. Since manual contour delineation is labor-intensive and time-consuming and suffers from inter-observer variability, it is highly desired in clinical studies to develop automatic medical image segmentation methods.

The advancements in deep learning have significantly leveraged the potential of deep neural networks in various medical image segmentation tasks \cite{bernard2018deep,heller2020state,lalande2021deep}.
Among various deep learning-based networks, U-Net \cite{3D-UNet,U-Net} and its variants \cite{li2018hdenseunet,zhang2020saunet,zhou2019unet++,yang2024d} are widely used and developed, following the encoder-decoder architecture.
Specifically, nnU-Net with automatically adapted training strategies and network architectures based on U-Net architecture has shown state-of-the-art performance on many medical image segmentation tasks \cite{nnunet}.
More recently, the success of transformer architectures in various computer vision tasks \cite{ViT2020} spurred the development of transformer-based segmentation methods, which have been increasingly applied in the field of medical image segmentation \cite{chen2021transu,xie2021cotr,zhou2023nnformer}.

\begin{figure*}[!t]
    \centering
    \includegraphics[width=18cm]{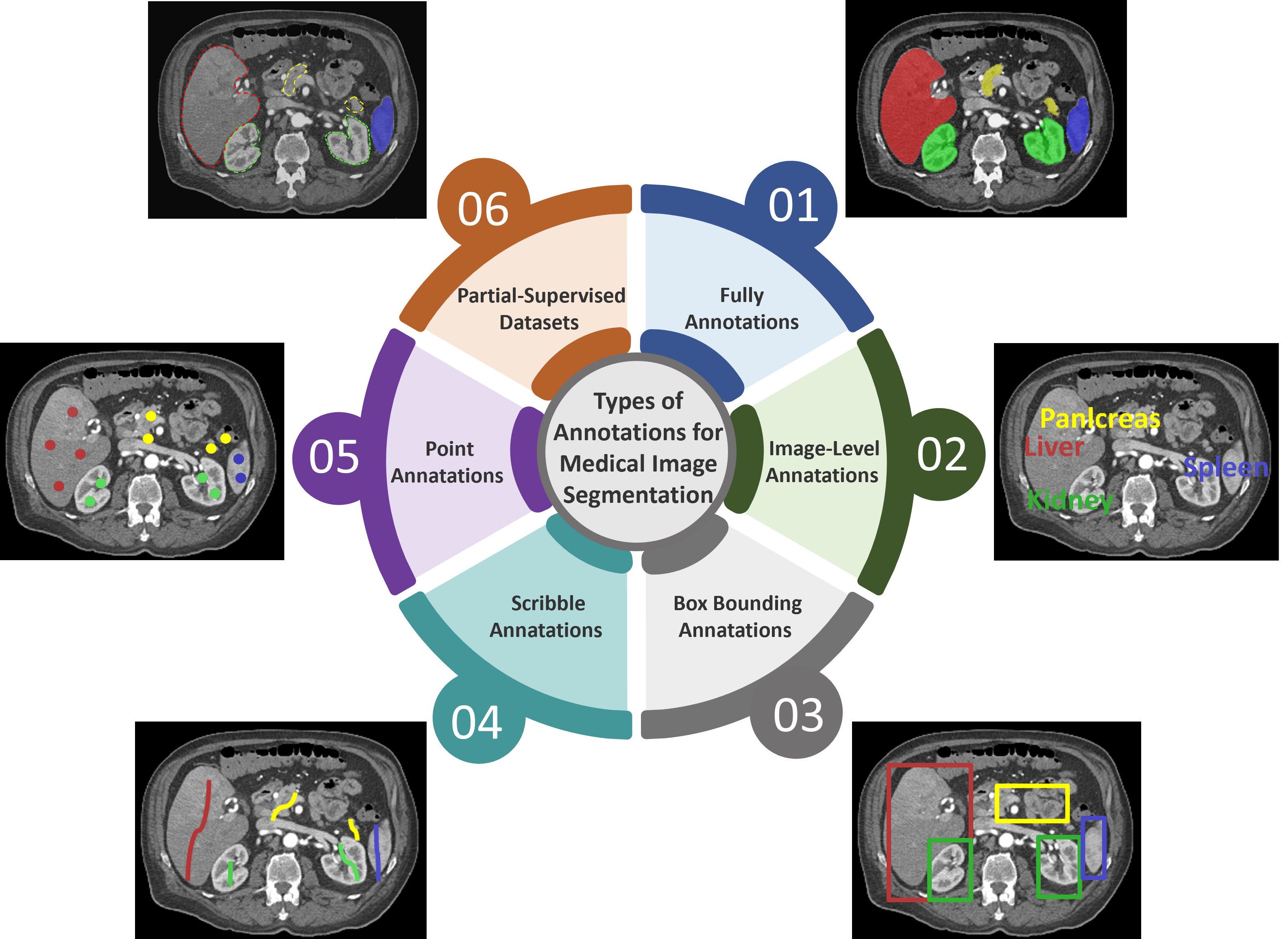}
    \caption{Example of different types of weak annotation compared with pixel-wise fully annotation for medical image segmentation tasks.}
    \label{Overview}
\end{figure*}

Although these architectural advancements have shown encouraging results, these cutting-edge segmentation methods mainly require large amounts of training data with pixel-wise annotations \cite{LabelEfficient}. However, manually annotating medical images at pixel-wise is a costly and time-consuming process, which necessitates the expertise of experienced clinical professionals. 
The scarcity of annotated medical imaging data is compounded by variations in patient populations, acquisition parameters, protocols, sequences, vendors, and centers, leading to significant statistical discrepancies \cite{campello2021multi}. Besides, many commonly used medical images like computed tomography (CT) and magnetic resonance imaging (MRI) are 3D volumetric data, which further increase the burden of manual annotation since experts need to delineate the target from the volume slice by slice \cite{2-5D-MIS}. The dense manual labeling can take several hours to annotate one image for experienced radiologists.
These challenges have stimulated extensive research into developing segmentation networks with limited annotations like semi-supervised learning \cite{jiao2023learning}. Despite the performance of these semi-supervised segmentation methods to utilize unlabeled data \cite{luo2022semi,zhang2023uncertainty,shi2023competitive}, these methods still require a subset of pixel-wise annotated data to achieve comparable performance. As an alternative, weak annotations like bounding boxes or scribbles can significantly reduce the annotation cost compared with pixel-wise fully annotations, as observed in Fig. \ref{Overview}.
Besides, the emergence of vision foundation models \cite{SAM} has also introduced innovative capabilities for segmentation tasks using weak annotations like points and bounding boxes for promptable segmentation enabled by large-scale pre-training.

In this paper, we present a comprehensive survey that encapsulates the evolution of weakly supervised medical image segmentation techniques, which have emerged as a response to the challenges associated with the high cost and expertise required for pixel-wise annotations in medical imaging.
By synthesizing the findings from various studies and presenting a unified perspective on weakly supervised medical image segmentation, this paper contributes to the collective understanding of the field and offers guidance for future research endeavors.

\begin{table*}
	\caption{The summarized review of weakly medical image segmentation methods with image-level annotations.}
	\centering
	\renewcommand\arraystretch{2.6}
        \scalebox{1.0}{
	\begin{tabular}{c|c|c|p{3.0cm}|p{7cm}}
	    \hline
        Reference &
          2D/3D &
          Modality &
          Dataset &
          Strategy to bridge the supervision gap \\
        \hline
        Jia \textit{et al.} \cite{jia2017constrained}   &
          2D &
          Microscope &
          TMAs \cite{xu2014weakly} &
          Incorporate multi-scale learning and area constraints for cancerous region segmentation. \\
        Feng \textit{et al.} \cite{feng2017discriminative} &
          3D &
          CT &
          LIDC-IDRI \cite{armato2011lung} &
         Learn discriminative regions from the activation maps of convolution units at different scales. \\
        Chen \textit{et al.} \cite{chen2022c} &
          3D &
          MRI &
          ProMRI \cite{litjens2014evaluation,clark2013cancer}, ACDC \cite{bernard2018deep}, CHAOS \cite{kavur2021chaos} &
          Leverage both category-causality and anatomy-causality to overcome ambiguous boundaries and co-occurrence. \\
		\hline
	\end{tabular}}
\end{table*}

\section{Task Formulation}

Given a defined set of $C$ classes represented by $\mathcal{C}=\{0,1,\dots,c-1\}$, for a given image (for 3D volumetric images) $\mathbf{X}\in\{\mathbb{R}\}^{H{\times}W{\times}D}$, the segmentation task aims to generate predictions $\mathbf{Y}\in\{\mathcal{C}\times\mathbb{N}\}^{H{\times}W{\times}D}$, where $\mathbf{y}_i=(c_i)\in\mathcal{C}\times\mathbb{N}$ of pixel/voxel $i$ represents the label tuple for the pixel/voxel in $\mathbf{X}$ at the same spatial location, and $H, W$ and $D$ are the height, width and depth of the image, respectively, $\mathbb{N}$ is the space of nature numbers, and $c_i$ represents the classification class of the pixel/voxel at spatial location $i$. 

For the training of segmentation models, given a training set $\mathcal{T}=\{(\mathbf{X}^{(n)},\mathbf{Y}^{(n)})\}$ with $N$ images, where $\mathbf{Y}^{(n)}\in\{\mathcal{C}\times\mathbf{N}\}^{H{\times}W{\times}D}$ represents the dense annotations of the image $\mathbf{X}^{(n)}$, which means that each pixel/voxel at spatial location $i$ of image $\mathbf{X}^{(n)}$ is annotated by a label tuple $\mathbf{y}^{(n)}_i$.
However, due to the manual labeling burden for pixel-wise annotations $\mathbf{Y}^{(n)}$ which is expensive and difficult to obtain, weakly supervised annotation-efficient methods focus on training segmentation models based on weak labels that cannot cover full supervision signals but are much easier to obtain. These annotation-efficient approaches are classified by the types of supervision from weak labels, which can be formulated from a unified perspective regarding the format of the training data. 
In the following, we introduce and give mathematical definitions for several commonly used weak annotations for medical image segmentation tasks.

\textbf{Image-level Annotation.} Image-level annotation is a type of annotation that utilizes only high-level image-level labels to train segmentation models. The training set can be formulated as $\mathcal{T}=\{(\mathbf{X}^{(n)},c^{(n)})|n\in\mathcal{N}\}$, where $c^{(n)}\subseteq\mathcal{C}$ is the image-level labels representing the presence or absence of each object class in the image.

\textbf{Bounding-box Annotation.} Bounding-box annotation aims to draw a rectangular box around the object of interest to indicate its location and size within the image. The training set can be formulated as $\mathcal{T}=\{(\mathbf{X}^{(n)},\mathcal{B}^{(n)})|n\in\mathcal{N}\}$ and $\mathcal{B}^{(n)}=\{(\mathbf{b}^{(n,m)},\mathbf{y}^{(n,m)})\}$, where $\mathbf{b}$ is the vertex coordinates of bounding box $\mathcal{B}$ of the image.

\textbf{Scribble/Point Annotation.} Sparse annotation represents that only a discriminative subset of the target is annotated. For example, scribble annotation is a set of scribbles with only coarse-grained or incomplete annotations where only rough object boundaries are provided. Specifically, each object is represented by a set of scribbles, which are manually drawn lines along or around the object's boundary. 
Point annotation aims to simply mark a few key points inside the object of interest in an image instead of drawing rough boundaries around it like in scribble annotation.
The training set can be formulated as $\mathcal{T}=\{(\mathbf{X}^{(n)},\bar{\mathbf{Y}}^{(n)})|n\in\mathcal{N}\}$, where $\bar{\mathbf{y}}^{(n)}\subseteq{\mathbf{y}}^{(n)}$ represents the sparse annotations including scribbles and points.

\textbf{Partially Supervision.}  In real-world applications, a great number of datasets are partially annotated to meet different practical usages, which is commonly seen in medical image analysis. For example, most medical datasets are collected for the segmentation of only one type of organ and/or tumors, while other task-irrelevant objects are treated as the background. Partially supervised segmentation tasks have been made to utilize multiple partially labeled datasets for multi-task segmentation. 
The training set can be formulated as $\mathcal{T}=\{(\mathbf{X}^{(n)},{\mathbf{Y}}^{(n)}_{c})|n\in\mathcal{N},c\in\mathcal{C}\}$, where ${\mathbf{y}}^{(n)}_{c}$ represents the partial annotations of class $c$ among all target classes $\mathcal{C}$.

\begin{figure}[!t]
    \centering
    \includegraphics[width=\linewidth]{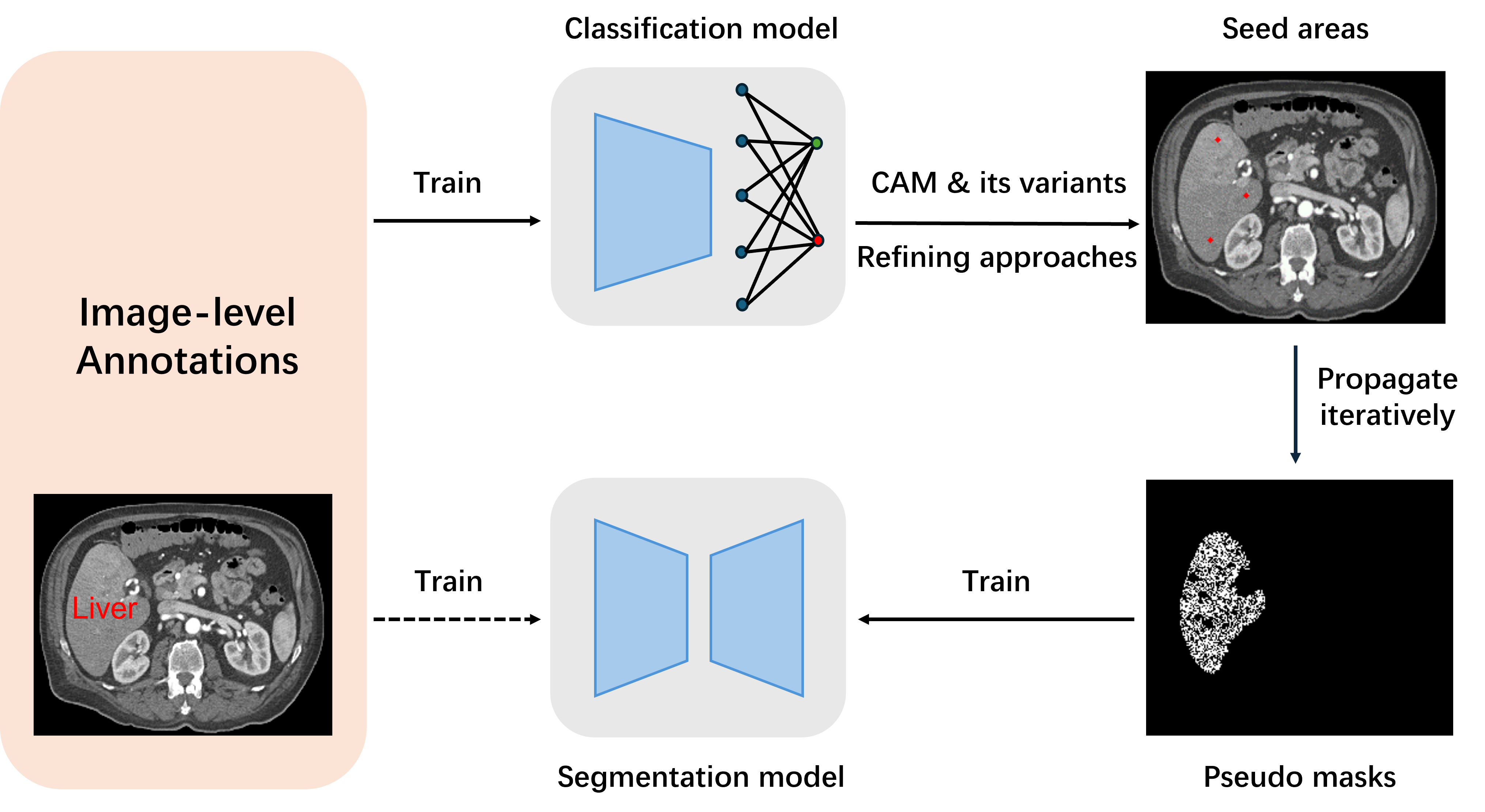}
    \caption{The overall workflow of weakly supervised medical image segmentation with image-level annotations.}
    \label{image-level}
\end{figure}

\begin{table*}
	\caption{The summarized review of weakly medical image segmentation methods with bounding boxes.}
	\centering
	\renewcommand\arraystretch{2.6}
        \scalebox{1.0}{
	\begin{tabular}{c|c|c|p{2.4cm}|p{7cm}}
	    \hline
        Reference &
          2D/3D &
          Modality &
          Dataset &
          Strategy to bridge the supervision gap \\
        \hline
        Rajchl \textit{et al.} \cite{rajchl2016deepcut}  &
          2D &
          MRI &
          IUGR dataset \cite{damodaram2012foetal} &
          Iterative energy minimization over densely-connected conditional random field. \\
        Wang \textit{et al.} \cite{wang2021bounding}  &
          3D &
          MRI &
          PROMISE12 \cite{litjens2014evaluation}, ATLAS \cite{liew2018large} &
          Integrate bounding box tightness prior with smooth maximum approximation. \\
        Redekop \textit{et al.} \cite{redekop2021medical} &
          3D &
          CT &
          MSD liver \cite{antonelli2021the} &
          Crop images within loose bounding boxes into patches for binary patch classification. \\
        Tang \textit{et al.} \cite{tang2021weakly} &
          3D &
          CT &
          NIH DeepLesion \cite{yan2018deeplesion} &
          Refine regions near segmentation boundary with level-set loss. \\
		\hline
	\end{tabular}}
\end{table*}

\section{Solutions for Medical Image Segmentation with Weak Annotations}

\subsection{Medical Image Segmentation with Image-Level Annotations}

Image-level annotation is one of the most efficient forms for weak-supervised learning, which provides only high-level class labels for each training image without any pixel-wise annotation. 
The major challenge of using image-level annotations for pixel-wise predictions lies in the significant disparity. Specifically, image-level annotations cannot provide sufficient information to guide models to accurately make pixel-wise predictions and outline interested targets.

An overall workflow of weakly supervised image segmentation with image-level annotations is illustrated in Fig. \ref{image-level}. In the first state, pseudo masks are generated for each training image based on a classification model trained with image-level supervision. After that, a segmentation model is trained based on the pseudo masks.
From the goal of applying the classification model is to generate high-quality pseudo masks from image-level annotations, which consists of two subsequent steps. Firstly, some seed areas are obtained in each training image based on the information derived from the classification model. Then the pseudo masks are generated by propagating the seed areas to the whole image. Since the pseudo masks are inevitably noisy due to the lack of supervision, the generation process is usually implemented in an iterative manner to enable the model to improve the quality of pseudo masks progressively.

For the generation of seed areas, most approaches are based on the concept of class activation mapping (CAM) \cite{CAM} and its variants \cite{GradCAM,GradCAM++,ScoreCAM} achieved by classification model that covers discriminative regions of the image. 
Specifically, Früh \textit{et al.} \cite{fruh2021weakly} evaluate the feasibility of weakly supervised framework by comparing different CAM methods including classic CAM \cite{CAM}, GradCAM \cite{GradCAM}, GradCAM++ \cite{GradCAM++} and ScoreCAM \cite{ScoreCAM} in tumor segmentation with image-level annotations. Experimental results demonstrate that CAM, GradCAM++, and ScoreCAM yielded similar results, while GradCAM led to inferior results that may ignore multiple instances in a given slice. 
Despite serving as an efficient tool to generate seed areas based on the classification model, CAM still suffers from several limitations. For example, CAM could only locate the discriminative part of the object and could not completely cover the object, which may deteriorate the performance.
Additionally, compared with natural images, medical images usually have low contrast resulting in objects have ambiguous boundaries. Therefore, directly applying CAM-based methods is usually less effective. Besides, the co-occurrence is very severe in medical images, which means that different segmentation targets (e.g. organs) often occur in the same image \cite{chen2022c}. As a result, it is hard to activate correct co-occurring targets in one image only according to image-level labels.

To address these challenges, several approaches are proposed in refining seed areas to make them more accurate and generate more reliable pseudo masks based on the seed areas.
Li \textit{et al.} \cite{li2022deep} propose a weakly-supervised framework with image-level annotations for breast tumor segmentation. Breast tumors are recognized by a classification model and then segmented based on class activation mapping and deep level set (CAM-DLS), where domain-specific anatomical information of breast ultrasound is utilized to reduce the search space.
Izadyyazdanabadi \textit{et al.} \cite{izadyyazdanabadi2018weakly} construct a weakly supervised framework to generate glioma Diagnostic Feature Maps (DFM) from confocal laser endomicroscopy (CLE) images using a multi-layer CAM followed by global average pooling to obtain the image-level label prediction. 
Hwang \textit{et al.} \cite{hwang2016self} present a self-transfer learning (STL) framework for weakly supervised lesion localization by jointly optimizing classification and localization networks to help the localization network focus on correct lesions without any types of priors.
Yoo \textit{et al.} \cite{yoo2022superpixel} propose a pipeline to use binary classification labels to segment regions of interest (ROIs) for brain tumor segmentation using MRI scans.
Feng \textit{et al.} \cite{feng2017discriminative} propose a two-stage weakly supervised framework that involves a coarse image segmentation followed by a fine-grained instance-level segmentation.  The first stage leverages CAM generated by an image classification model to determine the presence or absence of a nodule in each slice. Then, in the second stage, the region of interest around each localized instance in the class activation map is selected while everything outside this region is masked out. 
Jia \textit{et al.} \cite{jia2017constrained} propose a multiple instance learning framework for weakly supervised segmentation given only image-level cancer labels for histopathology images by introducing constraints about positive instances to effectively explore additional weakly supervised information. Each histopathology image is treated as a bag and each individual pixel in the image is treated as an instance. To regularize training, area constraints are further imposed to the segmentation results of multiple-instance learning.
Dubost \textit{et al.} \cite{dubost2020weakly} propose a new weakly supervised framework to compute attention maps that reveal the locations of brain lesions. using the last feature maps of a segmentation network optimized only with global image-level annotations.
Dang \textit{et al.} \cite{dang2022vessel} propose a novel annotation-efficient framework to utilize weak patch-level tag labels and a CAPTCHA-like setup to synthesize pixel-wise pseudo-labels for vessel and background.
Chen \textit{et al.} \cite{chen2022c} propose a causal class activation map (C-CAM) method for weakly-supervised medical image segmentation to generate pseudo segmentation masks with clearer boundaries and more accurate shapes by integrating two cause-effect chains including category-causality chain to alleviate the problem of ambiguous boundary and anatomy-causality chain to issue the co-occurrence challenge.

\begin{figure}[!t]
    \centering
    \includegraphics[width=\linewidth]{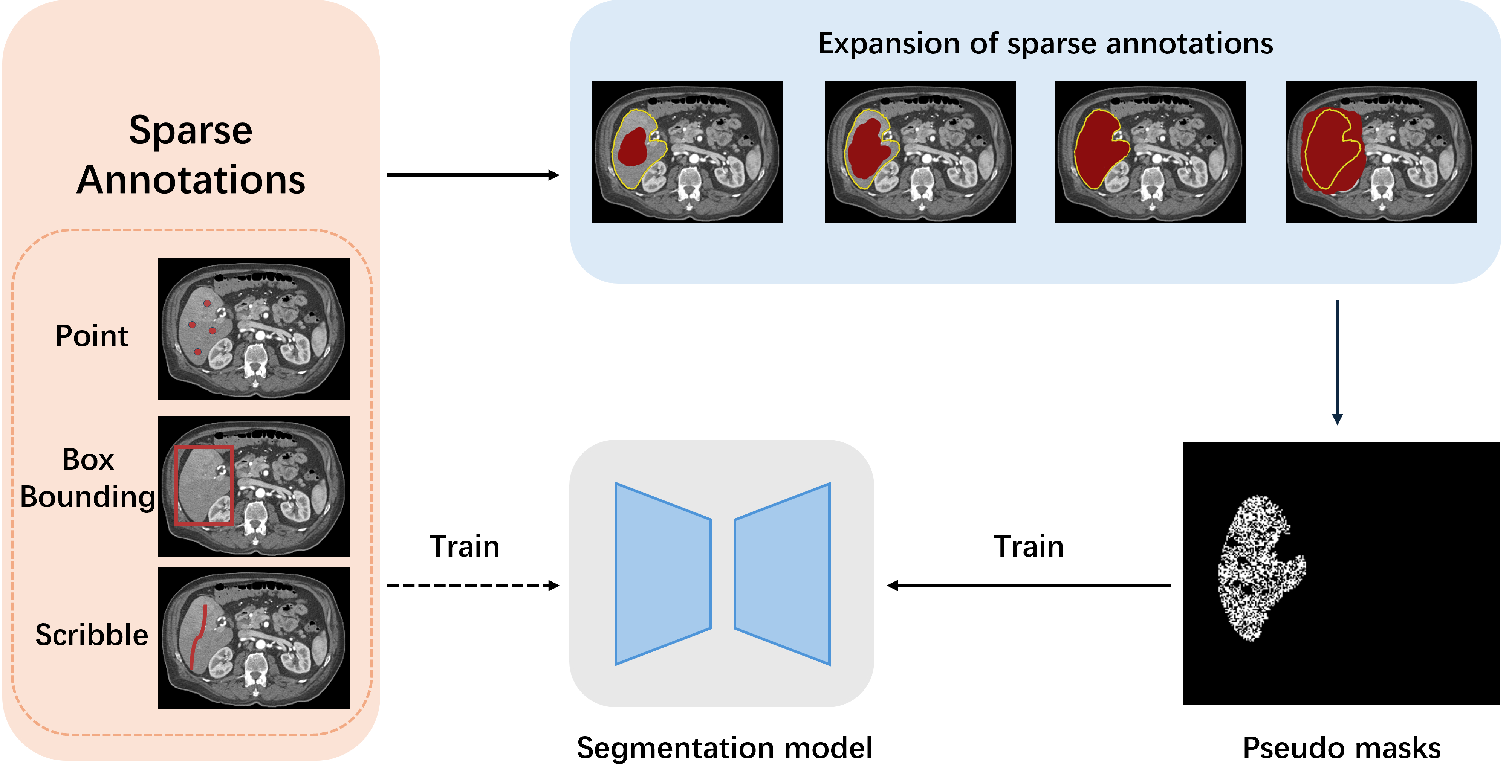}
    \caption{The overall workflow of weakly supervised medical image segmentation with sparse annotations like point, box bounding, and scribble.}
    \label{sparse}
\end{figure}

\begin{table*}
	\caption{The summarized review of weakly medical image segmentation methods with scribble annotations.}
	\centering
	\renewcommand\arraystretch{2.6}
        \scalebox{1.0}{
	\begin{tabular}{c|c|c|p{3.6cm}|p{7cm}}
	    \hline
        Reference &
          2D/3D &
          Modality &
          Dataset &
          Strategy to bridge the supervision gap \\
        \hline
        Zhang \textit{et al.} \cite{zhang2022cyclemix} &
          3D &
          MRI &
        ACDC \cite{bernard2018deep}, MSCMRseg \cite{zhuang2016multivariate, zhuang2018multivariate} &
        Maximize scribbles using mixed augmentation and random occlusion. \\
        Luo \textit{et al.} \cite{luo2022scribble} &
          3D &
          MRI &
          ACDC \cite{bernard2018deep} &
          Randomly mixing outputs of dual-branch network to generate pseudo labels. \\
        Asad \textit{et al.} \cite{asad2022ECONet}  &
          3D &
          CT &
          UESTC-COVID-19 \cite{wang2020noise} &
          Extract patches exclusively with scribbles for online likelihood learning. \\
        Yang \textit{et al.} \cite{yang2024non} &
          3D &
          MRI &
          CHAOS \cite{kavur2021chaos}, ACDC \cite{bernard2018deep}, LVSC  \cite{suinesiaputra2014collaborative}&
          Leverage a Siamese architecture with consistency training and entropy regularization. \\
        Zhang \textit{et al.} \cite{zhang2022shapepu} &
          3D &
          MRI &
          ACDC \cite{bernard2018deep}, MSCMRseg \cite{zhuang2016multivariate, zhuang2018multivariate}&
          Distinguish unlabeled pixels by maximizing the marginal probability. \\
        Li \textit{et al.} \cite{li2023scribblevc} &
          3D &
          MRI &
          ACDC  \cite{bernard2018deep}, MSCMRseg \cite{zhuang2016multivariate, zhuang2018multivariate}, NCI-ISBI  \cite{clark2013cancer} &
          Use gated conditional random field loss for emphasizing boundaries between regions. \\
		\hline
	\end{tabular}}
\end{table*}

\subsection{Medical Image Segmentation with Bounding Boxes}

Bounding-box annotations can provide valuable information to guide models to identify and localize regions of interest, so using bounding-box annotations for segmentation is a popular topic in the field of medical imaging.
Compared with image-level supervision, bounding box-level supervision serves is a more powerful form of annotation to narrow down the search space and enclose the segmented region within a rectangle, which offers a more comprehensive and informative representation of the target region's location and extent and enables more accurate and robust segmentation.

To utilize weak annotations, pseudo masks are generated based on the annotated bounding boxes as priors. After that, a segmentation model is trained based on the pseudo masks. The primary difficulty in weakly supervised medical image segmentation with bounding boxes lies in accurately distinguishing foreground objects from the background regions within the bounding box.
Rajchl \textit{et al.} \cite{rajchl2016deepcut} propose DeepCut to achieve weakly supervised medical image segmentation based on bounding box annotations. Specifically, the segmentation problem is formulated as an energy minimization problem over a densely connected conditional random field with an iterative optimization to obtain pixel-wise segmentation from MRI images. 
Wang \textit{et al.} \cite{wang2021bounding} introduce a method for weakly supervised image segmentation that leverages tight bounding box annotations and utilizes generalized multiple instance learning (MIL) and smooth maximum approximation to integrate the bounding box tightness prior into a deep neural network in an end-to-end manner.
Redekop \textit{et al.} \cite{redekop2021medical} develop a new bounding box correction framework to improve the tightness of non-tight 3D bounding box annotations used as weak labels, showing that this correction significantly enhances the performance of weakly supervised segmentation approaches.
Wang \textit{et al.} \cite{wang2020iterative} present a novel weakly supervised approach for accurate segmentation of the prostate and nearby organs in CT images by utilizing 3D bounding box annotations and incorporating a label denoising module into the iterative training scheme to gradually refine the segmentation results. 
Afshari \textit{et al.} \cite{afshari2019weakly} propose to utilize bounding boxes around tumor lesions for automatic tumor delineation in clinical PET scans by introducing a novel loss function that combines supervised and unsupervised components.
Liu \textit{et al.} \cite{liu2021automatic} propose a weakly supervised learning method to utilize bounding box annotations by generating pseudo masks and incorporating data pre-processing techniques for segmenting liver, spleen, and kidney from CT images.
Zhang \textit{et al.} \cite{zhang2021weakly} introduce a weakly-supervised teacher-student network (WSTS) that utilizes additional box-level-labeled data to accurately segment liver tumors in non-enhanced images.
Tang \textit{et al.} \cite{tang2021weakly} present a novel weakly-supervised universal lesion segmentation method based on HRNet \cite{HRNet} with a regional level set (RLS) loss for optimizing lesion boundary delineation and scale attention mechanisms to extract high-resolution deep image features crucial for accurate lesion segmentation. The RLS loss enables reliable and effective optimization in a weakly-supervised fashion, improving segmentation near lesion boundaries.




 
 


\begin{table*}
	\caption{The summarized review of weakly medical image segmentation methods with point annotations.}
	\centering
	\renewcommand\arraystretch{2.6}
        \scalebox{1.0}{
	\begin{tabular}{c|c|c|p{3.2cm}|p{7cm}}
	    \hline
        Reference &
          2D/3D &
          Modality &
          Dataset &
          Strategy to bridge the supervision gap \\
        \hline
        Kexrvadec \textit{et al.} \cite{kervadec2018constrained}  &
          3D &
          MRI &
            LV \cite{bernard2018deep}, VB \cite{bernard2018deep},  Prostate segmentation \cite{chu2015fully}&
            Enforce inequality constraints with differentiable penalties. \\
       Zhai \textit{et al.} \cite{zhai2023pa}  &
          2D 3D &
          MRI &
           VS \cite{shapey2019artificial}, BraTS2019 \cite{menze2014multimodal} &
        Contextual regularization using conditional random field and variance minimization and cross knowledge distillation. \\
        Buhmann \textit{et al.} \cite{buhmann2018synaptic} &
          3D &
          Microscope &
          CREMI \footnotemark[1]  &
          Compare predicted synaptic connections with real synaptic point annotations. \\
        Breznik \textit{et al.} \cite{breznik2023leveraging}  &
          2D &
        MRI &
          ACDC \cite{bernard2018deep}, POEM \cite{lind2013relationships} &
          Calculate boundary loss using distance maps derived from point annotations. \\
          
       Tian \textit{et al.} \cite{tian2020weakly}&
          2D &
          Microscope &
          MoNuSeg\cite{kumar2017dataset}, TNBC \cite{naylor2018segmentation}&
          Integrate sparse supervised learning from points to regions and impose contour-sensitive constraints. \\
		\hline
	\end{tabular}}
 \\
 \footnotetext[1]  01. https://cremi.org/
\end{table*}

\subsection{Medical Image Segmentation with Scribble Annotations}

Scribble is a powerful form of sparse annotation that can provide shape and size information for objects with complex shapes, as shown in Fig. \ref{sparse}. Scribble annotations are created by marking certain regions of an image with scribbles, which serve as guidance for the segmentation algorithm to accurately classify and delineate different objects or regions within the image. By requiring minimal input from users, scribble-based segmentation significantly reduces the manual effort required for detailed image labeling while still producing high-quality results. Consequently, this approach has practical applications in various fields such as medical imaging, scene understanding, object recognition, and video analysis. The strength of scribble segmentation lies in its ability to combine human intuition with computational power, resulting in more precise and efficient image segmentation.

Moreover, weakly supervised medical image segmentation based on scribble typically can involve leveraging the provided scribbles to guide the segmentation process. One common approach is to incorporate the scribble information into a loss function that encourages the segmentation model to produce predictions that align with the indicated rough outlines or hints. This can be achieved by leveraging the scribble annotations to set constraints for boundary areas in the loss function and guide models to calculate the unknown class directly. Asad \textit{et al.} \cite{asad2022ECONet} present an efficient convolutional neural network (CNN) approach and use weighted cross-entropy loss to address the class imbalance that may result from user interactions.  
Zhang \textit{et al.} \cite{zhang2022shapepu} propose ShapePU based on the Positive-Unlabeled (PU) learning framework and global consistency regularization, leveraging unlabeled pixels via PU learning and exploiting shape knowledge. 
Can \textit{et al.} \cite{can2018learning} explore training strategies for learning the parameters of a pixel-wise segmentation network solely from scribble annotations.
Dorent \textit{et al.} \cite{dorent2020scribble} point out that scribbles on the target domain are used to perform domain adaptation with a new formulation of domain adaptation based on structured learning and co-segmentation.
Wang \textit{et al.} \cite{wang2018interactive} enhance the framework by integrating CNNs into a pipeline for bounding box and scribble-based segmentation and adjusting parameters to make a CNN model adaptable to a specific test image.

Another approach is expanding scribble into the complete annotation, which utilizes foreground masking techniques to indicate or employs a background mask to indicate, which areas belong to the foregrounds.
Lin \textit{et al.} \cite{lin2016scribblesup} show a weakly-supervised method for semantic segmentation based on scribbles, optimizing a graphical model for propagating information from scribbles. 
Xu \textit{et al.} \cite{xu2021scribble} reveal a Progressive Segmentation Inference (PSI) framework to tackle scribble-supervised semantic segmentation, which encapsulates two crucial cues, contextual pattern propagation, and semantic label diffusion. 
Luo \textit{et al.} \cite{luo2022scribble} utilize a dual-branch network comprising a single encoder and two marginally distinct decoders for image segmentation and blend the predictions from both decoders to create pseudo labels for auxiliary supervision. 
Zhang \textit{et al.} \cite{zhang2022cyclemix} propose a novel weakly supervised segmentation framework, CycleMix that combines mix augmentation and cycle consistency. It adopts the mixup strategy with a specially designed random occlusion to perform increments and decrements of scribbles. 
Zhang \textit{et al.} \cite{zhang2023zscribbleseg} exhibit to utilize the sole scribble annotations, investigating the principle of ”good scribble annotations”, which leads to efficient scribble forms via supervision maximization and randomness simulation, introducing regularization terms to encode the spatial relationship and shape prior, integrating the efficient scribble supervision with the prior into a unified framework.
Li \textit{et al.} \cite{li2023scribblevc} introduce a novel framework for
scribble-supervised medical image segmentation that leverages vision and class embeddings via the multimodal information enhancement mechanism, utilizing both CNN features and transformer features uniformly to achieve enhanced visual feature extraction.
Yang \textit{et al.} \cite{yang2024non} present the non-interactive method named PacingPseudo by comparing two weight-sharing networks, designing entropy regularization, distorted augmentations, and a new memory bank mechanism that provides an extra source of ensemble features to complement scarce labeled pixels.  
Zhuang \textit{et al.} \cite{zhuang2023annotation} give a training method only scribble guidance in the difficult areas. It uses a small set of fully annotated data to train the segmentation network and generates pseudo labels. Human supervisors draw scribbles in the areas of incorrect pseudo labels, and the scribbles are converted into pseudo label maps using a probability-modulated geodesic transform.
Lee \textit{et al.} \cite{lee2020scribble2label} introduce Scribble2Label only hands-less scribble annotations and combines pseudo-labeling and label filtering to generate reliable labels from weak supervision.

\subsection{Medical Image Segmentation with Point Annotations}

Point annotations are created by marking specific points of interest or landmarks on an image. These points are represented by coordinates $(x, y)$ indicating their positions within the image, as shown in Fig. \ref{sparse}. This annotation technique is commonly used for tasks like object detection, key point localization, facial landmark detection, and pose estimation.
Point annotations can provide precise information about the locations of important features or objects, enabling algorithms to make accurate predictions and perform detailed analysis.
When dense pixel-wise annotations are lack, weakly supervised segmentation with point annotation can be an effective method that uses auxiliary information or heuristic methods to estimate object or feature locations for annotation. Common methods include region-level annotation, weak label annotation, and auxiliary information annotation. 

The region-level annotation work mainly involves drawing bounding boxes and assigning labels. Zhai \textit{et al.} \cite{PASeg} propose a two-stage weakly supervised learning framework called PA-Seg for annotating segmentation in 3D medical images. The annotator only needs to provide seven points: one inside the target object as a foreground seed and six outside the target as background seeds.
Breznik \textit{et al.} \cite{breznik2023leveraging} investigate the combination of intensity-based distance maps with boundary loss for point-supervised semantic segmentation, where the boundary loss penalizes false positives farther away from the object more strongly.
The weak label annotation method uses fuzzy or partially labeled data for annotation.
Yoo \textit{et al.} \cite{yoo2019pseudoedgenet} introduce PseudoedgeNet, a weakly supervised nuclei segmentation method that only requires point annotations for training.
Gao \textit{et al.} \cite{gao2020renal} manifest a framework that employs Minimal Point-Based annotation to accurately detect cancerous regions. The annotator only needs to mark a few cancerous and non-cancerous points in each whole-slide image (WSI).
Buhmann \textit{et al.} \cite{buhmann2018synaptic} propose a 3D U-Net architecture to identify pairs of voxels that are previous and postsynaptic to each other, formulating the problem of synaptic partner identification and allowing to directly learn from synaptic point annotations.
Due to annotation uncertainty, results may introduce some noise or error. Therefore, caution is advised when using weakly supervised point annotation for training and analysis, considering the accuracy and reliability of the results.

In addition, several methods have been proposed as cascaded frameworks or coarse-to-fine frameworks to address the problem of limited or sparse annotations. Qu \textit{et al.} \cite{qu2020weakly} propose a weakly supervised nuclei segmentation framework that utilizes partial points annotation to train a detection model in the first stage and a segmentation model in the second stage, achieving competitive performance with significantly less annotation effort compared to fully supervised methods. Kexrvadec \textit{et al.} \cite{kervadec2018constrained} propose a weakly supervised segmentation framework that uses inequality constraints with a differentiable term to enforce constraints directly in the loss function, which has the potential to bridge the gap between weakly and fully supervised learning.
Zhai \textit{et al.} \cite{zhai2023pa} indicate a method to annotate a segmentation target with only seven points, using geodesic distance transform in the first stage and leveraging model predictions as pseudo labels in the second stage.
Tian \textit{et al.} \cite{tian2020weakly} express a coarse-to-fine weakly-supervised framework that employs self-stimulated learning through a self-supervision strategy using clustering for binary classification. 
Roth \textit{et al.} \cite{roth2019weakly,roth2021going} propose to speed up medical image annotation by using minimal user interaction in the form of extreme point clicks to train a segmentation model, which is refined through multiple rounds of training and custom-designed loss and attention mechanism.

\begin{table*}
	\caption{The summarized review of weakly medical image segmentation methods with partially-supervised datasets.}
	\centering
	\renewcommand\arraystretch{2.6}
        \scalebox{1.0}{
	\begin{tabular}{c|c|c|p{3.2cm}|p{7cm}}
	    \hline
        Reference &
          2D/3D &
          Modality &
          Dataset &
          Strategy to bridge the supervision gap \\
        \hline
        Zhang \textit{et al.} \cite{zhang2021dodnet} &
          3D &
          CT &
          LiTS \cite{bilic2023liver}, KiTS \cite{heller2020state}, MSD \cite{antonelli2021the}  &
          Use a single network with a dynamic segmentation head to train from partially labeled datasets.\\
        Dmitriev \textit{et al.} \cite{dmitriev2019learning} &
          3D &
          CT &
          Sliver07 \cite{heimann2009comparison}, NIH Pancreas \cite{holger2016turkbey}&
          Incorporate conditional information to implicitly share all parameters among target classes. \\
        Zhou \textit{et al.} \cite{zhou2019prior} &
          3D &
          CT &
          BTCV \cite{landman2015miccai} &
          Incorporate anatomical priors to approximate empirical organ size distributions. \\
		\hline
	\end{tabular}}
\end{table*}

\begin{figure}[!t]
    \centering
    \includegraphics[width=\linewidth]{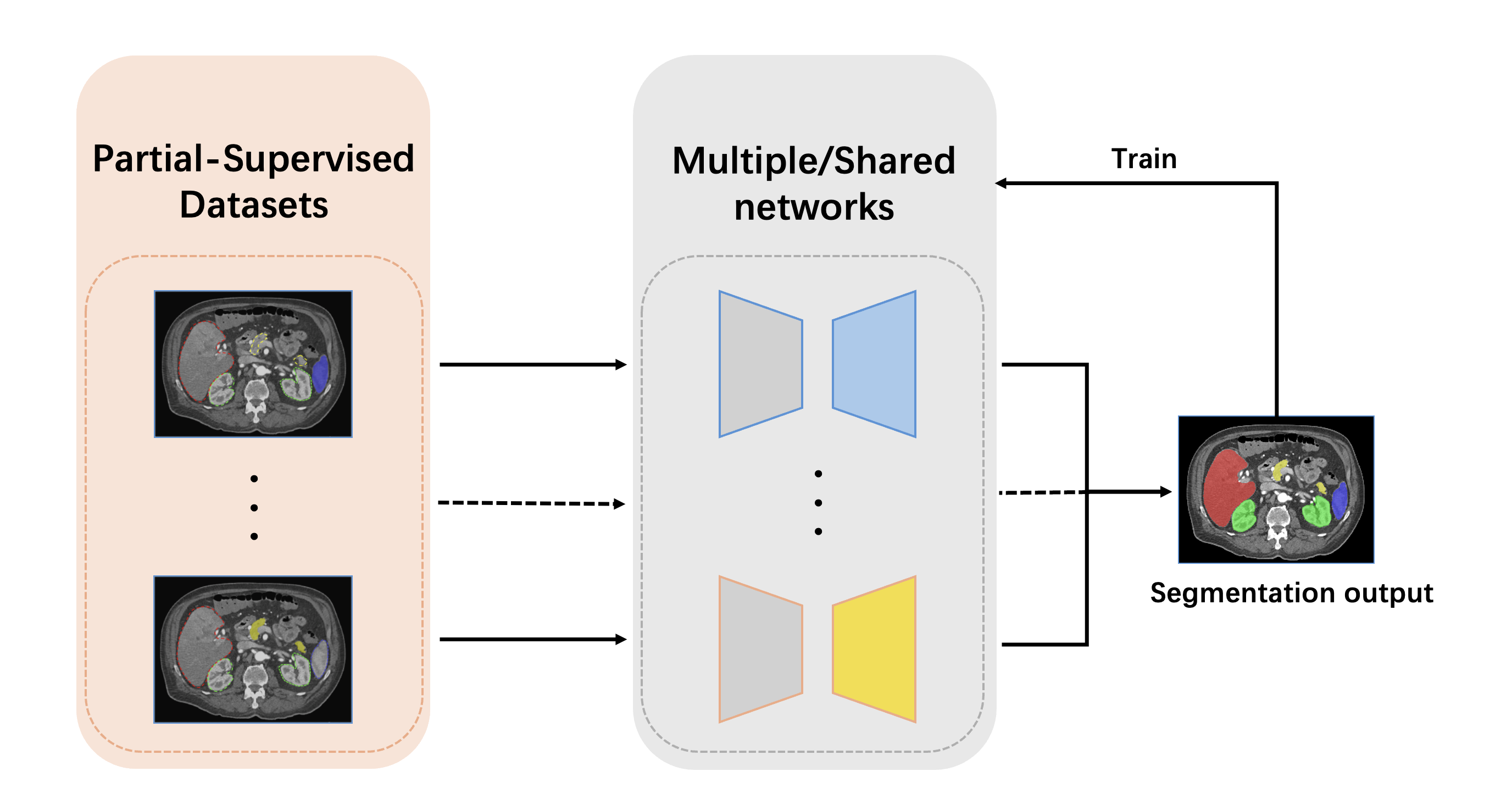}
    \caption{The overall workflow of weakly supervised medical image segmentation with partially-supervised datasets.}
    \label{partial}
\end{figure}

\subsection{Medical Image Segmentation with Partially-Supervised Datasets}

The workflow for medical image segmentation with partially-supervised datasets involves several steps. Firstly, the medical images and their corresponding annotations are collected and preprocessed. Then, the dataset is split into a fully-supervised subset and a partially-supervised subset. Next, a fully-supervised segmentation model is trained using the fully-supervised subset. Using this model, pseudo labels are generated for the partially-supervised subset. Subsequently, a segmentation model is trained on the combined fully-supervised and pseudo-labeled partially-supervised subsets, employing weakly-supervised or semi-supervised methods. To further enhance segmentation accuracy, the model is fine-tuned using the fully-supervised subset. Finally, the model is evaluated on a separate test set to validate its performance. This workflow enables accurate medical image segmentation even with limited labeled data by leveraging both fully and partially supervised learning techniques.

Zhang \textit{et al.} \cite{zhang2021dodnet} propose a dynamic on demand-network (DoDNet) which learns to segment multiple organs and tumors on partially labeled datasets. DoDNet consists of a shared encoder-decoder architecture, a task encoding module, a controller for dynamic filter generation, and a single but dynamic segmentation head. The task's information from current segmentation tasks is encoded as a task-aware prior, guiding the model on the expected goals and outcomes of the task.
Dmitriev \textit{et al.} \cite{dmitriev2019learning} present a unified highly efficient framework for robust simultaneous learning of multi-class segmentation by combining single-class datasets and utilizing a novel way of conditioning a convolutional network for segmentation.
Zhou \textit{et al.} \cite{zhou2019prior} propose Prior-aware Neural Network (PaNN) to address the background ambiguity in partially-labeled datasets, which incorporates anatomical priors on abdominal organ sizes and guides the training process with domain-specific knowledge.
Fidon \textit{et al.} \cite{fidon2021partial} advocate the first axiomatic definition of label-set loss functions which are the loss functions that can handle partially segmented images, proving that there is one and only one method to convert a classical loss function for fully segmented images into a proper label-set loss function. 
Huang \textit{et al.} \cite{huang2020partly} design a novel general-purpose semi-supervised, multiple-task model—namely, self-supervised, semi-supervised, multitask learning (S4MTL) in medical imaging, segmentation, and diagnostic classification.

\subsection{Medical Image Segmentation with Mixed Supervision}

Mix supervised learning in weakly supervised medical image segmentation addresses the problem of limited or incomplete annotations by combining different types of supervision signals. It leverages both weak annotations, such as scribbles or points, and strong annotations, such as fully segmented images, to improve the accuracy and robustness of the segmentation model. By combining these different sources of supervision, mixed supervised learning allows the model to learn from both coarse-level guidance and fine-level details, resulting in more accurate and detailed segmentations. This approach helps mitigate the challenges associated with weakly supervised segmentation and enhances the overall performance of the segmentation model in medical imaging tasks. This strategy is particularly important in cases where high-quality labeled data is scarce or expensive to obtain because it allows models to learn from weaker forms of supervision and still achieve high accuracy. Overall, mixed supervision has significant implications for the advancement of machine learning and artificial intelligence by enabling more efficient and effective learning from diverse data sources.

Mix supervised learning in weakly supervised medical image segmentation involves collecting a dataset with both weakly annotated images and strongly annotated images, designing a neural network architecture capable of handling both types of supervision signals, and training the network using a combination of weak and strong annotations. The loss function typically includes both weak and strong supervision terms, encouraging the network to learn from coarse-level guidance and fine-level details. By leveraging multiple sources of supervision, mix supervised learning improves the accuracy and robustness of medical image segmentation models, enabling the network to produce more precise and reliable segmentations. To address the challenge of limited annotated datasets for deep neural networks (DNNs) in medical image segmentation, Liu \textit{et al.} \cite{liu2023segmentation} proposes a dual-branch architecture with a novel formulation incorporating Shannon entropy loss and Kullback-Leibler divergence to leverage mixed supervision, achieving substantial performance improvements compared to other strategies and recent semi-supervised approaches.

In order to address the high demand for accurate annotations in pixel-wise segmentation tasks, Rei \textit{et al.} \cite{rei2021every} present a semi-weakly supervised segmentation algorithm that utilizes deep supervision and a student-teacher model, achieving significant reduction in the requirement for expensive labels and narrowing the performance gap to fully supervised approaches using different types of annotations.
Utilizing deep supervision and a student-teacher model allows for easy integration of different supervision signals. By carefully integrating deep supervision in lower layers and employing multi-label deep supervision, the authors achieve significant improvements in segmentation performance. Through a novel training regime that effectively uses various levels of annotation. Sun \textit{et al.} \cite{sun2020teacher} develop a semi-supervised learning framework based on a teacher-student fashion, propose a hierarchical organ-to-lesion (O2L) attention module in a teacher segmentor to produce pseudo-labels and train a student segmentor with combinations of manual- labeled and pseudo-labeled annotations.

By leveraging both weak and strong annotations and designing an appropriate loss function, mixed supervised learning enables the network to learn from different levels of supervision and improve the accuracy and robustness of medical image segmentation models. 
Shah \textit{et al.} \cite{shah2018ms} propose a new FCN named MS-Net to reduce supervision cost by coupling strong supervision with weak supervision through low-cost input in the form of bounding boxes and landmarks.
Shen \textit{et al.} \cite{shen2019simultaneous} give a mixed-supervision guided method and a residual-aided classification U-Net model (ResCU-Net) for joint segmentation and benign-malignant classification, which coupling the strong supervision in the form of segmentation mask and weak supervision in the form of benign-malignant label.
Dolz \textit{et al.} \cite{dolz2021teach} indicate a dual-branch architecture. The upper branch teacher receives strong annotations but the bottom branch student is driven by limited supervision and guided by the upper branch, encouraging confident student predictions at the bottom branch and transferring the knowledge from the predictions generated by the strongly supervised branch to the less-supervised branch.
Upadhyay \textit{et al.} \cite{upadhyay2019mixed} show a novel generative adversarial gan network that can leverage training data at multiple levels of quality to improve performance while limiting the costs of data acquisition.
Bhalgat \textit{et al.} \cite{bhalgat2018annotation} present a budget-based cost-minimization framework in a mixed-supervision setting via dense segmentation, bounding boxes, and landmarks. The linear programming combines uncertainty with a similarity-based ranking strategy to select annotated examples.

\section{Revisiting Weak Annotations in the Era of Foundation Models}

The emergence of large-scale foundation models \cite{wang2023large,liang2022foundations} has revolutionized artificial intelligence (AI) and sparked a new era, primarily due to their remarkable zero-shot and few-shot generalization abilities across a wide range of downstream tasks \cite{bommasani2021opportunities}. 
Foundation models not only hold significant potential in natural language processing but have also broadened their application scope into computer vision with the emergence of vision-based foundation models \cite{wang2023seggpt,wang2023mis}.

One pioneer example of vision foundation models \cite{jung2023uncover} is the Segment Anything Model (SAM) \cite{kirillov2023segment}, which has gained significant attention due to its impressive performance on a variety of semantic segmentation tasks \cite{semanticSAM,trackanything}. The design of SAM focuses on versatility, operational efficiency, and the inherent ability to handle ambiguity, representing significant progress in this field. It integrates an image encoder, a prompt encoder, and a mask decoder to generate masks for all objects within an image, demonstrating unprecedented zero-shot generalization for unseen objects or tasks. SAM's versatility is evident in its ability to handle various types of prompts  \cite{jung2023uncover} such as points, bounding boxes, and free-form text, making it adaptable for various segmentation tasks. This adaptability is achieved through a combination of positional encodings for points and bounding boxes \cite{tancik2020fourier} and the complex integration of different embeddings for each type of cue, thereby enhancing its ability to effectively interpret various segmentation tasks. Significantly, the architecture of SAM allows it to produce multiple masks for a single prompt, effectively handling ambiguous cases \cite{kirillov2023segment}. This finding points to a significant aspect of vision foundation models: the utilization of weak annotations as prompts, which is fundamentally driven by the need for models to efficiently handle ambiguity and generalize from limited information. By effectively processing these simple prompts, SAM can adapt to various tasks and scenarios with increased efficiency, showing its versatility in diverse applications. This strategy represents a significant shift towards more adaptable and resource-efficient AI models.

In the domain of medical image segmentation with weak annotations, the application of SAM signifies a pivotal shift towards more efficient and scalable methodologies \cite{SAM4MIS}. This shift is evidenced by a collection of studies that aim to optimize the annotation process and enhance the performance of segmentation models, leveraging minimal yet effective inputs like bounding boxes and point prompts. The evolution of these methods showcases a concerted effort to address the inherent challenges of medical image segmentation.
Zhang \textit{et al}. \cite{zhang2023efficiently} indicate that using SAM with bounding box annotations to train segmentation models has high cost-effectiveness. Compared to models trained with pixel-wise annotations, this method shows competitive segmentation performance without the high demand for fully supervised annotations. This approach can provide a scalable and effective approach for medical image segmentation, potentially revolutionizing the annotation process and model training in this field.
Building on this foundation, Deng \textit{et al}. \cite{deng2023segment} further evaluate SAM's performance in a more focused context, specifically in segmenting tumors, non-tumor tissues, and cell nuclei from whole slide imaging (WSI) utilizing both point annotations (both single and multiple) and bounding box annotations for annotation. They identify SAM's remarkable performance in segmenting large connected objects but also uncover limitations in dense instance segmentation, pointing to the need for model fine-tuning and improved prompt selection.

To adapt to the nuanced demands of specific segmentation tasks and the challenge of effectively utilizing weak annotations, several studies have introduced tailored modifications to SAM, indicating its versatility and robustness in addressing these specialized needs.
Feng \textit{et al}. \cite{feng2023cheap} employ image synthesis from a few exemplars to augment the dataset, and Low-Rank Adaptation (LoRA) \cite{hu2021lora} for fine-tuning SAM. This method demonstrates promising results in segmenting medical images with weak annotations, particularly in tasks like brain tumor segmentation and multi-organ CT segmentation, without specifying the exact type of weak annotations used for initial exemplar generation.
In addition, Cui \textit{et al}. \cite{cui2023all} propose a pipeline that utilizes SAM, called all-in-SAM, for cell nucleus segmentation from bounding boxes to pixel-level labels,  significantly exceeding the results obtained by supervised learning methods. This indicates that SAM has a strong ability to handle weak annotations of bounding boxes, and can clearly detect the nuclear boundaries within the focus area.
Similarly, Wang \textit{et al}. \cite{wang2023sam} introduce $\text{SAM}^{\text{Med}}$, a framework for medical image annotation using SAM, which consists of $\text{SAM}^{\text{assist}}$ and $\text{SAM}^{\text{auto}}$. Among them, $\text{SAM}^{\text{assist}}$ not only significantly improves the accuracy of SAM for medical image segmentation with only about five input points, but also combines points and bounding boxes to simulate real-world annotation scenes, demonstrating the adaptability of SAM.

Since SAM was originally designed for 2D natural images, it faces limitations in effectively extracting 3D spatial information from volumetric medical data \cite{awais2023foundational}. 
To address these limitations of SAM, Wu \textit{et al}. \cite{wu2023medical} propose a novel framework called Medical SAM Adapter (Med-SA) to enhance SAM's ability to use weak annotations for high-quality medical image segmentation. Med-SA significantly expands SAM's applicability in various medical imaging modes by introducing Space-Depth Transpose (SD-Trans) and Hyper-Prompting Adapter (HyP-Adpt). SD-Trans applies 2D SAM to 3D medical images, while HyP-Adpt allows prompt conditional model adaptation, enabling the model to effectively utilize click prompts and bounding boxes to generate accurate segmentation masks.
Moreover, Gong \textit{et al}. \cite{gong20233dsam} also develop a novel method, 3DSAM-adapter, to extend the application of SAM from 2D to 3D images in medical data. This method adapts to 3D medical data by modifying the image encoder, prompt encoder, and mask decoder, effectively utilizing spatial context to improve segmentation performance. This adaptive improvement demonstrates superiority on four common tumor segmentation datasets using single-point annotations per volume, highlighting the potential and efficiency of using vision foundation models to handle weak annotations in the field of medical image segmentation.

SAM modes have been further improved by incorporating auxiliary modules or other software to capture additional contextual information.
Liu \textit{et al}. \cite{liu2023samm} describe the Segment Any Medical Model, an extension of SAM integrated into 3D Slicer \cite{fedorov20123d} for medical image segmentation with weak annotations.
This method involves the use of prompt points as annotations, where users interactively place points on the image slices to indicate regions of interest or to exclude certain areas. These prompt points are then used by the model to generate segmentation masks.
Similarly, Lei \textit{et al}. \cite{lei2023medlsam} develop MedLSAM, a model for automatic medical image segmentation that integrates a few-shot localization model \cite{lei2021contrastive} with the segmentation prowess of SAM for medical image segmentation. The localization model identifies anatomical regions across CT scans using self-supervision, reducing reliance on extensive manual annotations. It generates bounding boxes for image slices, which SAM utilizes for precise segmentation. This approach leverages weak annotations, such as minimal bounding boxes, simplifying segmentation tasks and enhancing the efficiency of processing limited labeled medical imagery. Zhang \textit{et al}. \cite{UR-SAM} further adopt prompt augmentation of generated bounding box prompts for uncertainty estimation and rectification to enhance the reliability of segmentation results.
Shahabany \textit{et al}. \cite{shaharabany2023autosam} present AutoSAM, an improved SAM model specifically designed for medical image segmentation tasks. AutoSAM incorporates an auxiliary prompt encoder to directly use input images as prompts, rather than relying on manual weak annotations that SAM originally needed. This method enables AutoSAM to automatically adapt to the characteristics of medical images, improve segmentation performance, and reduce dependence on complex manual annotations. AutoSAM expands the application scope of SAM, making it more suitable for medical image segmentation tasks while maintaining the characteristics of weakly supervised learning.
Chen \textit{et al}. \cite{chen2023weakly} explore the further applicability of SAM within the realm of weakly supervised segmentation by integrating text inputs and zero-shot learning settings. In the text input scenario, SAM employs class labels as text prompts, in combination with Grounded DINO \cite{liu2023grounding} to generate bounding boxes, which are then used to produce segmentation masks. In the zero-shot learning scenario, the Recognize Anything Model (RAM) \cite{zhang2023recognize} is leveraged to identify class labels in the absence of explicit annotations, followed by a similar process of generating bounding boxes and masks. This method promotes medical image segmentation by explaining different prompts and demonstrating the potential of basic models in effectively processing weakly annotated data.

In addition to SAM and its variants, several researches also focus on other visual foundation models to expand the scope of medical image segmentation. 
Li \textit{et al}. \cite{li2023promise} propose ProMISe, a 3D medical image segmentation model using only a single point prompt. ProMISe utilizes point annotation to enable the model to focus on specific regions, improving accuracy and robustness in complex medical image segmentation tasks.
Du \textit{et al}. \cite{du2023segvol} introduce SegVol, a universal and interactive volumetric medical image segmentation foundation model. SegVol employs semantic and spatial prompts, including image-level annotations, bounding boxes, and point annotations, for precise segmentation through weak annotations. This model combines pseudo labels generated from unlabeled data with weakly supervised learning from a small amount of labeled datasets. Numerous experiments have demonstrated that SegVol outperforms state-of-the-art models on multiple segmentation benchmarks, particularly emphasizing its effectiveness in lesion segmentation on challenging datasets.

Collectively, these studies underscore the paradigm shift towards leveraging weak annotations and pioneering adaptations of vision foundation models for medical image segmentation. They highlight the continuous innovation in this field, enhancing the efficiency and applicability of segmentation models. This body of work paves the way for more scalable, annotation-effective solutions in medical imaging, promising significant advancements in the accuracy and efficiency of medical diagnoses and treatments.

\section{Discussion and Conclusion}

The rapid advancements in the field of medical image segmentation, particularly with the advent of deep learning techniques, have significantly improved the accuracy and efficiency of automatically delineating anatomical structures or lesions from medical images. To ease the heavy reliance on pixel-wise annotations, which are labor-intensive and require expert knowledge, weakly supervised learning and the emergence of foundation models have opened new avenues for training deep models with limited or imperfect annotations, thus addressing the challenges associated with manual annotation.
The weak supervision strategies have demonstrated the potential to reduce the reliance on extensive manual annotations while still achieving competitive segmentation performance. These methods leverage different forms of weak labels to guide the learning process, from high-level image-level labels to coarse scribbles and point locations, enabling the models to learn from available, yet limited weak annotations.

By combining the analysis of the latest literature, the following challenges are worthy of further discussion.

\textbf{1) Quality evaluation and control of weak annotations.}
Assessing the quality of weak annotations is a critical challenge in weakly-supervised medical image segmentation. The performance of the segmentation model is heavily dependent on the quality of the weak annotations, which lack the granularity and precision compared with pixel-wise annotations \cite{zhang2023zscribbleseg}. 
Moreover, weak annotations come in various forms, and the disparity in quality across these different types can significantly impact the model's performance. Weak annotations may originate from multiple sources including different annotators, annotation tools, or guidelines, and may contain noise, such as incorrect labels or inaccurate bounding boxes. These ambiguities may burden the development and lead to indecisiveness in the model during segmentation.

To address this challenge, there are several possible solutions.
Firstly, the integration of various types of weak annotations, or combining weakly supervised learning with semi-supervised learning can enhance the contextual information for the model. Besides, it is important to develop specific metrics to quantify annotation quality, such as assessing consistency among different annotations or calculating overlap with a known "gold standard" \cite{hartmann2022mism}. Furthermore, incorporating probabilistic models and graph learning techniques can be beneficial, as they can account for uncertainties in annotations and better capture the complex relationships within medical images.

\textbf{2) Integrating domain knowledge.} For medical image segmentation tasks, some organs have intricate structures that are difficult to segment accurately without pixel-wise supervision. For example, intra-kidney variability requires precise delineation that weakly-supervised methods may struggle to achieve due to the lack of fine-grained annotations \cite{du2023weakly}.
Integrating domain knowledge can be a possible solution to overcome data scarcity and enhance the generalization capabilities and interpretability of segmentation model. For image data, one way is to explore prior information like position constraints and anatomical priors.

\textbf{3) Utilizing existing datasets.}
Utilizing existing external datasets can help improve model generalization ability, which may come from different medical laboratories or related diseases \cite{zhang2021exploiting}. These data may contain diverse information and enable the model to learn more knowledge specific to the medical field.
The recent introduction of vision foundation models like the Segment Anything Model \cite{SAM} has also shown the ability empowered by large-scale pre-training from natural images, with application to medical image segmentation tasks with weak annotations \cite{zhao2023segment}. These models have also expanded the scope of medical image segmentation by incorporating weak annotations more effectively.

In conclusion, the survey presented in the document highlights the significant progress in medical image segmentation with weak annotations. The shift from traditional models to foundation models marks a paradigm change towards more flexible and scalable approaches that can generalize from limited information. The weakly supervised learning techniques have proven to be effective in reducing the annotation burden while maintaining high segmentation accuracy, which is crucial for clinical applications where expert time is valuable and resources are limited.
The future of medical image segmentation lies in the continued development and refinement of these weakly supervised methods, as well as the integration of foundation models that can leverage weak annotations for improved performance. With the ongoing advancements in artificial general intelligence, it is expected that these models will become increasingly adept at handling ambiguous and limited data, further revolutionizing the field of medical imaging and contributing to more accurate diagnoses and effective treatments.
As the research community continues to explore the capabilities of foundation models and weakly supervised learning, it is also crucial to focus on the practical applications and the potential impact on healthcare. The ultimate goal is to develop models that can seamlessly integrate into clinical workflows, providing radiologists and medical professionals with powerful tools to enhance the quality and speed of medical diagnostics. The path forward involves collaborative efforts between researchers, clinicians, and AI developers to ensure that these technologies are not only technologically advanced but also practically relevant and beneficial to the medical community.

\bibliographystyle{IEEEtran}
\bibliography{ref}

\end{document}